\documentclass[conference]{IEEEtran}
\IEEEoverridecommandlockouts
\usepackage{cite}
\usepackage{amsmath,amssymb,amsfonts}
\usepackage{algorithmic}
\usepackage{overpic}
\usepackage{graphicx}
\usepackage{subcaption}
\usepackage{textcomp}
\usepackage{amsmath}
\usepackage{amsthm}
\usepackage{multirow} 
\usepackage{adjustbox}
\usepackage{makecell}
\usepackage{xcolor}
\def\BibTeX{{\rm B\kern-.05em{\sc i\kern-.025em b}\kern-.08em
    T\kern-.1667em\lower.7ex\hbox{E}\kern-.125emX}}
\begin{document}

\title{AnyArtisticGlyph: Multilingual Controllable Artistic Glyph Generation\\

\author{
    \IEEEauthorblockN{Xiongbo Lu$^{1}$, Yaxiong Chen$^{2,1,3}$, Shengwu Xiong$^{2,1,3}$\thanks{* Corresponding Author}}    
    \IEEEauthorblockA{$^1$ School of Computer Science and Artificial Intelligence, Wuhan University of Technology, Wuhan 430070, China}
}

}

\maketitle

\begin{abstract}
    Artistic Glyph Image Generation (AGIG) differs from current creativity-focused generation models by offering finely controllable deterministic generation. It transfers the style of a reference image to a source while preserving its content. Although advanced and promising, current methods may reveal flaws when scrutinizing synthesized image details, often producing blurred or incorrect textures, posing a significant challenge. Hence, we introduce AnyArtisticGlyph, a diffusion-based, multilingual controllable artistic glyph generation model. It includes a font fusion and embedding module, which generates latent features for detailed structure creation, and a vision-text fusion and embedding module that uses the CLIP model to encode references and blends them with transformation caption embeddings for seamless global image generation. Moreover, we incorporate a coarse-grained feature-level loss to enhance generation accuracy. Experiments show that it produces natural, detailed artistic glyph images with state-of-the-art performance. Our project will be open-sourced on https://github.com/jiean001/AnyArtisticGlyph to advance text generation technology.
\end{abstract}


\section{Introduction} \label{sec:intro}
Due to the excellent fidelity, adaptability, and versatility of diffusion-based generative models \cite{zhang2023adding}, they can provide certain creative inspiration for human design and other work. In contrast to the purpose of these methods, Artistic Glyph Image Generation (AGIG) is more like a finely controllable deterministic generation, which is a task that involves transferring the reference image style onto the source while preserving its original content. It can greatly reduce the labor of artistic font design due to the complex combination of lines, serif details, colors, and textures, particularly for logographic languages like Chinese (with over 60K characters), Japanese (over 50K characters), and Korean (over 11K characters). Therefore, AGIG has wide-ranging applications, such as creating new artistic fonts, restoring ancient characters, and augmenting data for optical character recognition.
\begin{figure}
    \centering 
    \begin{overpic}[width=\columnwidth]{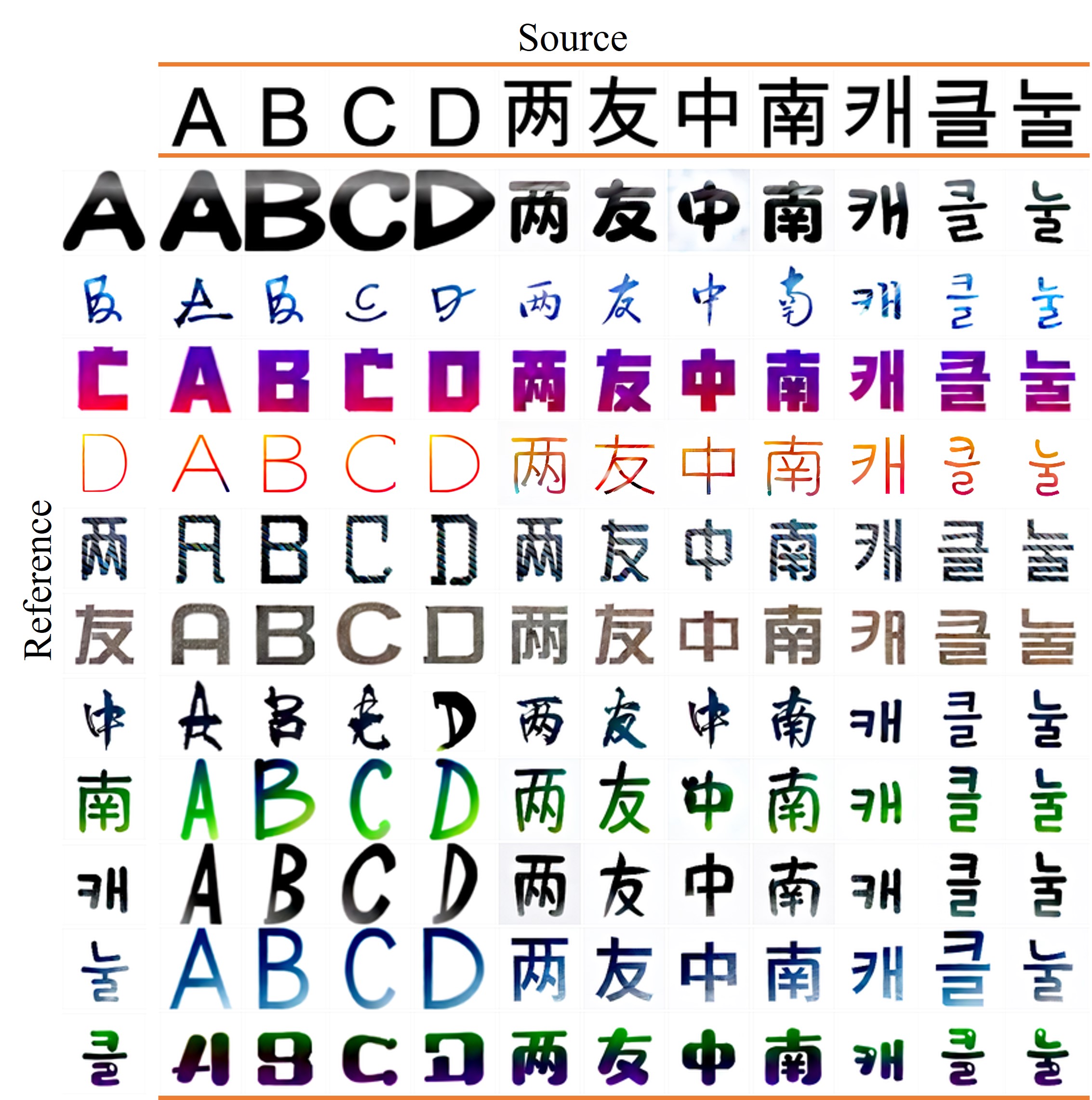}
    \end{overpic}
    \caption{Selected samples of AnyArtisticGlyph for cross-linguistic glyph generation.}
    \label{fig-intro-compare}
    \vspace{-15pt}
 \end{figure}

Recently, numerous methods have emerged for generating glyph and artistic glyph images, which have shown promising outcomes. However, these methods can continue to be improved from various perspectives to improve generation performance. Firstly, the current AGIG is commonly defined as the task of Image-to-Image translation (\cite{Isola_2017_CVPR}) to learn the transformation information implicit in the image. While this information can be provided by other modalities such as text. Secondly, due to the large appearance differences between characters in different languages, these methods usually focus on solving the generation of a specific language, e.g. \cite{azadi2018multi} employs stack-based input for Latin generation. Lastly, as pointed out in \cite{Fontdiffuser}, some methods design certain structure-based or component-based loss functions for generation, which hinders the practical application of such methods. 

To address the aforementioned difficulties, we present the AnyArtisticGlyph framework and A$^2$Glyph-24 dataset. AnyArtisticGlyph consists of a diffusion pipeline with two primary elements: a Font Fusion and Embedding Module (FFEM) employs inputs like text, and reference artistic glyph images to generate latent features for the final generation, especially the detailed structure generation. A Vision-Text Fusion and Embedding Module (VTFEM) use the CLIP model for encoding reference data as embeddings, which blend with explicit transformation caption embeddings from the stable diffusion to accomplish seamless global image generation. Moreover, a coarse-grained feature-level loss for training to enhance generation accuracy further prevents the model from solely emphasizing pixel-level features. 

% \begin{table}[]
%    \centering
%    \caption{Functionality Comparison: AnyArtisticGlyph vs. Diffusion-Based Competitors.}
%        \begin{tabular}{c|ccc}
%        \Xhline{2pt}
%        Functionality  & Cross-lingual   & Fancy-style  & Complex-structure   \\
%        \Xhline{1pt}
%        Diff-Font \cite{Diff_Font}        & $\times$      & $\times$    & $\times$                \\
%        FontDiffuser \cite{Fontdiffuser}  & \checkmark    & $\times$    & \checkmark              \\
%        \Xhline{1pt}
%        AnyArtisticGlyph     & \checkmark  & \checkmark  & \checkmark          \\
%        \Xhline{2pt}
%        \end{tabular}
%        \label{tab-intro}
%        \vspace{-10pt}
% \end{table}
% Regarding the functionality, three differentiating factors set apart us from other competitors as outlined in Tab. \ref{tab-intro}: a) Cross-lingual: AnyArtisticGlyph can be applied to the cross-lingual glyph generation tasks such as Chinese to Korean, showcasing the cross-domain generalization ability. b) Fancy-style: it enables the generating of different colors, textures, and large style variation glyph images. c) Complex structure: our method can accurately and precisely generate complex local and global features. We randomly present some cross-linguistic examples in Fig. \ref{fig-intro-compare}.

\begin{figure*}
    \centering 
    \begin{overpic}[width=1.98\columnwidth]{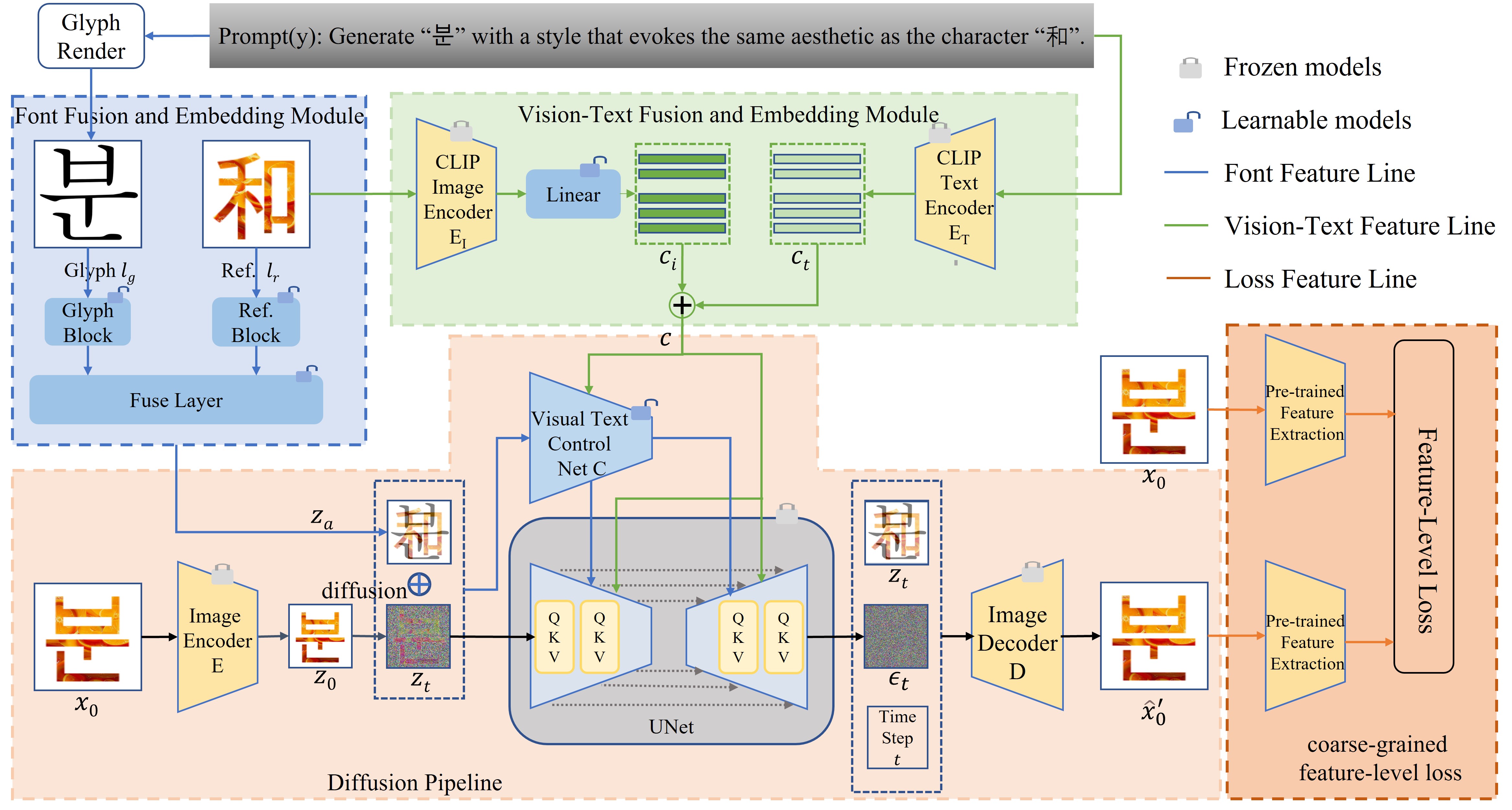}
    \end{overpic}
    \caption{The framework of AnyArtisticGlyph, which includes a diffusion pipeline, font fusion and embedding module, vision-text fusion and embedding module, and coarse-grained feature-level loss.}
    \label{fig-method-model}
    \vspace{-10pt}
 \end{figure*}

 The main contributions of our work can be summarized as follows:
 \begin{itemize}
 \item We propose a unified framework named AnyArtisticGlyph that supports multi-language text generation and cross-language synthesis. 

 \item We introduce the FFEM, which generates latent features for detailed structure creation using text and reference images.

 \item We develop the VTFEM that uses the CLIP model to encode reference data and blend it with transformation caption embedding for seamless global image generation.

 \item Comprehensive experiments on multilingual datasets demonstrate the effectiveness of the proposed AnyArtisticGlyph.
 \end{itemize}

\section{RELATED WORKS}
\textbf{Artistic glyph image generation} can be broadly categorized into font synthesis and style transfer. Font generation, focusing on shape extraction and creation, represents a specialized grayscale-level application within this domain. Early research \cite{zi2zi, chang2018chinese} approached artistic glyph generation as an image-to-image translation problem, facing difficulties in generating novel styles. To address these challenges, researchers proposed disentanglement \cite{sun2017learning, zhang2018separating} and unsupervised methods \cite{FUNIT, DGFont}. For specific languages, certain methods \cite{park2021multiple, kong2022look, liu2022xmp} leveraged structured components to capture local features, but labeling these components was labor-intensive and impractical for many languages. To overcome this, prior-free methods \cite{NTF, Fontdiffuser, wang2023cf, tang2022few} were introduced, such as \cite{Fontdiffuser}, which modeled AGIG as a noise-to-denoise task, incorporating a Multi-scale Content Aggregation block to preserve intricate strokes. However, these methods still rely heavily on image content, limiting the generation of complex and stylistically diverse glyphs

\textbf{Diffusion Model.} Since the introduction of the Diffusion Model (DM) by Jascha et al. \cite{Jascha2015}, it has undergone rapid advancements across diverse generation tasks. To enhance controllability and practicality, notable conditional diffusion models have been proposed, including those in \cite{zhang2023adding, ruiz2023dreambooth, Rombach_2022_CVPR}. Notably, LDM \cite{Rombach_2022_CVPR} facilitates DM training on limited computational resources through a latent space denoising process, maintaining image quality and flexibility. This foundation has led to the commercial application of the SD big model \cite{borji2022generated}. In text image generation, DM has demonstrated promising results in generating handwritten characters \cite{gui2023zero}. Related works, such as Diff-Font \cite{Diff_Font} and FontDiffuser \cite{Fontdiffuser}, while image-driven, may produce artifacts. Unlike general image generation, AGIG necessitates heightened control, accuracy, and predictability. Given the impressive results of current open-source diffusion-based models in content, detail, and structural features, we aim to leverage SD's capabilities and propose an innovative fine-tuning strategy tailored for stylistic glyph generation.

\section{METHOD} \label{method}
As depicted in Fig. \ref{fig-method-model}, the AnyArtisticGlyph framework comprises a pre-trained open-source text-control diffusion pipeline with two primary components. The overall training objective is defined as:
\abovedisplayskip=3pt
\begin{equation}
    L = L_{df} + \lambda L_{fl}.
    \label{eq-loss-total}
\end{equation}
\belowdisplayskip=5pt
where $L_{df}$ and $L_{fl}$ are traditional diffusion loss and coarse-grained feature-level loss, and $\lambda$ is the hyper-parameter to adjust the weight ratio between these two loss functions. 

\subsection{Diffusion Pipeline}
In the diffusion pipeline, we input the target image $x_0 \in \mathbb{R}^{H \times W \times 3}$ to the image encoder $E$ to obtain its latent representation $z_0 \in \mathbb{R}^{h \times w \times c}$. Where c represents the latent feature dimension and $h \times w$ for the feature resolution downsampled by a factor. Subsequently, t-step noise progressively diffuses to $z_0$, resulting in the noisy latent image $z_t$. Here, $t \in (0, T)$ is the randomly sampled diffusion step, and $T$ is the pre-defined maximum diffusion step. Then, given a set of conditions including font feature $z_a \in \mathbb{R}^{h \times w \times c}$ by FFEM, visual-text embedding $c$ by VTFEM, as well as sampled time step t, a pre-trained open source diffusion model applies a noise predictor P having UNet structure to predict the noise $\hat{\epsilon_t}$ contained in $z_t$ with the following objective:
\begin{equation}
    L_{df} = \mathbb{E}_{z_{t}, z_{a}, c, t, \epsilon \in N(0, 1)}[||\epsilon - P(z_{t}, z_{a}, c, t)||_2^2],
    \label{eq-loss-df}
\end{equation}
where $L_{df}$ is the traditional diffusion loss. 

\subsection{Font Fusion and Embedding Module}\label{sec_ALM}
Our proposed method uses glyph image $l_g$ and the given reference image $l_r$ to generate font latent feature map $z_a$. For more details, glyph $l_g$ is produced by rendering characters utilizing a uniform font style (i.e., "Arial Unicode") onto an image based on the given prompt $y$. The rendered image $l_g$ can be seen here as the source image provided in the form of an image in the traditional AGIG method. This design comes at almost no additional cost. At the same time, the form of images it provides is more standardized and unified, and it is more practical. On the other hand, text-driven style character generation can explicitly provide the model with specific transformations from one character to another.

We use glyph block to downsample the glyph image $l_g$ and reference block to downsample the reference image $l_r$ for incorporating the image-based conditions. The glyph block G and reference block R both contain several Resnet stacked convolutional layers to capture the local and font-based features. After transforming these image-based conditions into feature maps that match the latent space of diffusion, we utilize a convolutional fusion layer F to merge $l_g$ and $l_r$, resulting in a generated feature map denoted as $z_a$, which can be represented as:
\begin{equation}
    z_a = F(G(l_g) \oplus R(l_r)),
    \label{eq-cal-z_a}
\end{equation}
where $\oplus$ represents concatenation, $z_a$ shares the same number of channels as $z_t$,.

\subsection{Vision-Text Fusion and Embedding Module}
Based on the pure image-based control in Section \ref{sec_ALM}, this paper also adds text control to enhance the performance of final style character generation. Traditional AGIG is more like an image-to-image translation problem, and introducing text information can guide the model to generate more understandable content. In AnyArtisticGlyph, we introduce a novel approach to address the problem of multilingual and cross-lingual glyph generation. Specifically, the reference image $l_r$ through frozen CLIP Image Encoder $E_I$ and trainable Linear $F_I$ to produce the global reference condition $c_i$ as follows:
\begin{equation}
    c_i = F_I(E_I(l_r)).
    \label{eq-ci}
\end{equation}
On another hand, the prompt $y$ is used to freeze pre-trained open source diffusion CLIP Text Encoder $E_T$ to capture the transfer condition $c_t$ as follows:
\begin{equation}
    c_t = E_T(y).
    \label{eq-ct}
\end{equation}
These two conditions $c_i$ and $c_t$ are then added (e.g., $c=c_i+c_t$) to get a fused intermediate representation, which will then be mapped to the intermediate layers of the UNet using a cross-attention mechanism. Due to the utilization of reference images for the condition instead of relying solely on language-specific text encoders, our approach significantly enhances the generation of multilingual and cross-lingual glyph images.

\subsection{Coarse-Grained Feature-Level Loss}
This paper introduces the feature-level loss $L_{fl}$ to encourage the model to prioritize high-level features over pixel differences. Additionally, since the diffusion model requires gradual denoising to produce the image, this process remains time-consuming. Therefore, we employ a feature-level loss on the coarse-grained generative image. The coarse-grained generated image $\hat{x}_0$ is obtained by one-step denoising from the predicted noise $\hat{\epsilon_t}$ as follows.
\begin{equation}
    \hat{x}_0 = D(\frac{z_t - \sqrt{1 - \bar{\alpha}_t}\textbf{I}\hat{\epsilon}_t}{\bar{\alpha}_t}),
\label{eq-pred_y}
\end{equation}
where $D$ is the image decoder as shown in Fig \ref{fig-method-model}, $\textbf{I}$ is the identity matrix, $\bar{\alpha}_t$ is the time-dependent diffusion coefficient.

Finally, the definition  of $L_{fl}$ as follows.
\begin{equation}
    L_{fl} = \bar{\alpha}_t(L_c(\hat{x}_0, l_g) + L_s(\hat{x}_0, l_r))
\label{eq-loss-fl}
\end{equation}
where $L_c(\cdot, \cdot)$ and $L_s(\cdot, \cdot)$ respectively represent content loss and style loss.

\begin{figure}[htbp] 
   \centering 
   \begin{minipage}{\columnwidth} 
       \centering  
       \begin{subfigure}{\columnwidth}
           \includegraphics[width=\columnwidth]{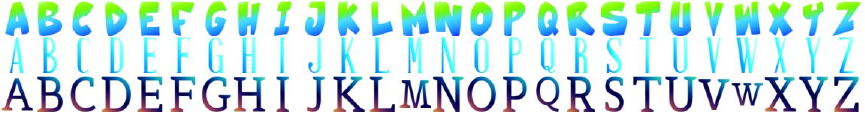} 
           \caption{MCGAN-Dataset artistic glyph images}  
           \vspace{3pt}
           \label{fig-data-english-style}  
       \end{subfigure}  
       \begin{subfigure}{\columnwidth} 
           \includegraphics[width=\columnwidth]{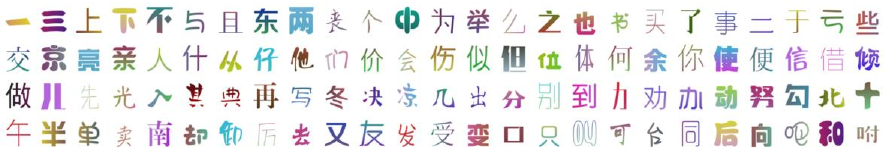}  
           \caption{Chinese100-Dataset artistic glyph images}  
           \label{fig-data-chinese-style} 
       \end{subfigure} 
       \begin{subfigure}{\columnwidth} 
           \includegraphics[width=\columnwidth]{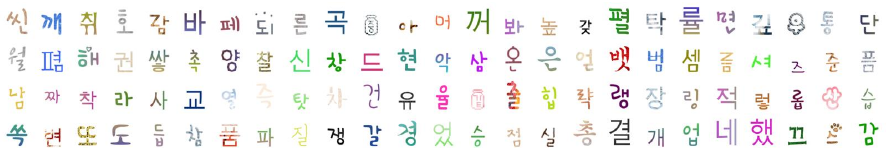}  
           \caption{Korean480-Dataset artistic glyph images}  
           \label{fig-data-Korean-style} 
       \end{subfigure}  
   \end{minipage}  
   \vspace{-3pt}
   \caption{Examples of synthetic artistic glyph images.}  
   \vspace{-15pt}
   \label{fig-data-style}  
\end{figure} 

\section{EXPERIMENTS}
\subsection{Experimental Setup}
\textbf{Dataset.} Currently, there is a lack of publicly available datasets specifically tailored for text-driven artistic glyph image generation tasks. Therefore, we propose A$^2$Glyph-24 dataset, a large-scale multilingual dataset from publicly available and synthesized images. The sources of these images including: \textbf{MCGAN-Dataset} is the English dataset proposed by MCGAN \cite{azadi2018multi}, which contains 32046 synthetic artistic fonts, each with 26 glyphs as shown in Fig. \ref{fig-data-english-style}; This paper builds the Chinese stylistic glyph image dataset named \textbf{Chinese100-Dataset}, which is constructed following the data processing methods in CF-Font \cite{CFFont} and MCGAN \cite{azadi2018multi} to expand the Chinese dataset. The Chinese100-Dataset comprises 9900 different styles, including 6900 training sets and 3000 test sets, with each style containing 100 Chinese glyphs. As depicted in Fig. \ref{fig-data-chinese-style}, these styles encompass various character glyphs, colors, textures, etc., ensuring diversity and an adequate representation of Chinese characters. Furthermore, to comprehensively evaluate the versatility and generalization of the method, this paper also builds a new Korean stylistic glyph image dataset named \textbf{Korean480-Dataset} based on the same method. The Korean480-Dataset comprises 10000 different styles, including 7000 training sets and 3000 test sets, with each style containing 480 Korean glyphs, some of them as shown in Fig. \ref{fig-data-Korean-style}.

\textbf{Benchmark}. For quantitative evaluation, we employ five widely used metrics in various image generation tasks: L1 loss, Structural Similarity (SSIM) index, Peak Signal-to-Noise Ratio (PSNR), Fréchet Inception Distance (FID), and Learned Perceptual Image Patch Similarity (LPIPS) index. Among them, L1, SSIM, and PSNR are pixel-level metrics calculated without the need for a pre-trained feature extractor, whereas FID and LPIPS operate at the perceptual level.

\textbf{Competitors}. We compare our AnyArtisticGlyph with the seven state-of-the-art image-driven methods: MCGAN \cite{azadi2018multi}, FET \cite{FET_GAN}, FUNIT \cite{FUNIT}, DG-Font \cite{DGFont}, NTF \cite{NTF}, Diff-Font \cite{Diff_Font} and FontDiffuser \cite{Fontdiffuser}. Among these, MC-GAN can only handle 26 English capital letters and is utilized solely for the English generation. Both Diff-Font and FontDiffuser are diffusion-based, while the rest are GAN-based methods. For the fairness of the experiment, each comparison method employs one-shot driven generation without fine-tuning the model. Additionally, the given content of each image-driven method remains consistent, and all methods are trained based on their official codes.

\begin{figure}
    \centering
    \begin{overpic}[width=\columnwidth]{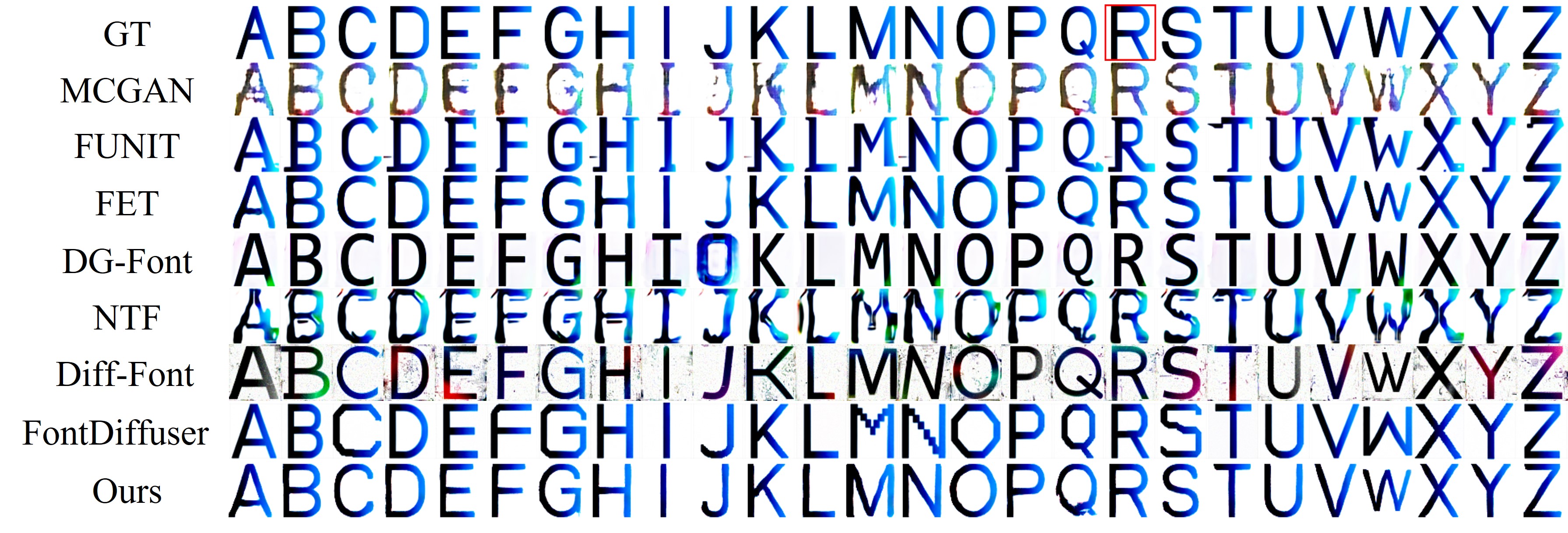}
    \put(0, 33.5){\scriptsize }
    \end{overpic}
    \caption{Comparison between competitors and our AnyArtisticGlyph on MCGAN-Dataset, where the ground truth is in the 1st row and the reference image with the red squares.}
    \label{exp_mcgan}
    \vspace{-10pt}
 \end{figure}

\textbf{Implementation Details}. Our training framework is implemented based on ControlNet\footnote{https://github.com/lllyasviel/ControlNet}, and the weights of AnyArtisticGlyph are initialized from SD1.5\footnote{https://huggingface.co/runwayml/stable-diffusion-v1-5}. In the experiments, Adam is used to optimize the model for 30 epochs, and the learning rate is initialized to 1e-5 and a batch size of 16. The hyperparameters in Equ. \ref{eq-loss-total} is fixed at $\lambda = 1$. For a fair comparison, image dimensions of $l_g$ and $l_r$ are set to $128 \times 128$, while $c_i$ and $c_t$ are all set to $77 \times 768$. The experimental results we report are the average of 3 tests. We implement our network using PyTorch and train it on CPU: E5-2609V4*2 @1.70GHz; GPU: Nvidia GeForce RTX 3090 GPU; RAM: 128GB; Ubuntu 20.04.
\begin{table}[]
    \centering
    \caption{Quantitative Results on MCGAN-Dataset. The bold indicates the state-of-the-art and the underline indicates the second best.}
    \begin{tabular}{c|cccccc}
    \Xhline{2pt}
    \multirow{2}{*}{Model} & \multicolumn{5}{c}{Metrics} \\
     \cline{2-6}
            & L1↓    & SSIM↑  & PSNR↑  & FID↓   & LPIPS↓        \\
    \Xhline{1pt}
    MCGAN   & 0.1915    & 0.4698    & 9.2766    & 53.3290   & 0.3319          \\
    FUNIT   & 0.2046    & 0.4458    & 8.3148    & 32.7251   & 0.3019          \\
    FET     & 0.1781    & 0.4971    & 8.9742    & 27.4484   & 0.2675          \\
    DG-Font & 0.2499    & 0.3740    & 7.1546    & 70.8665   & 0.3984          \\
    NTF     & 0.1831    & 0.4604    & 8.9042    & 53.0411   & 0.3057          \\
    Diff-Font  & 0.2984 & 0.1377    & 6.8416    & 109.1884  & 0.4714          \\
    FontDiffuser & \underline{0.1769}  & \textbf{0.5211}  & \underline{9.5143}    & \underline{19.6057}   & \underline{0.2442}          \\
    Ours    & \textbf{0.1755}     & \underline{0.5096}    & \textbf{10.2209}    & \textbf{11.1322}   & \textbf{0.2237}          \\
    \Xhline{2pt}
    \end{tabular}
    \label{tab-exp-mcgan}
 \end{table}

\begin{figure*}
   \begin{overpic}[width=1.95\columnwidth]{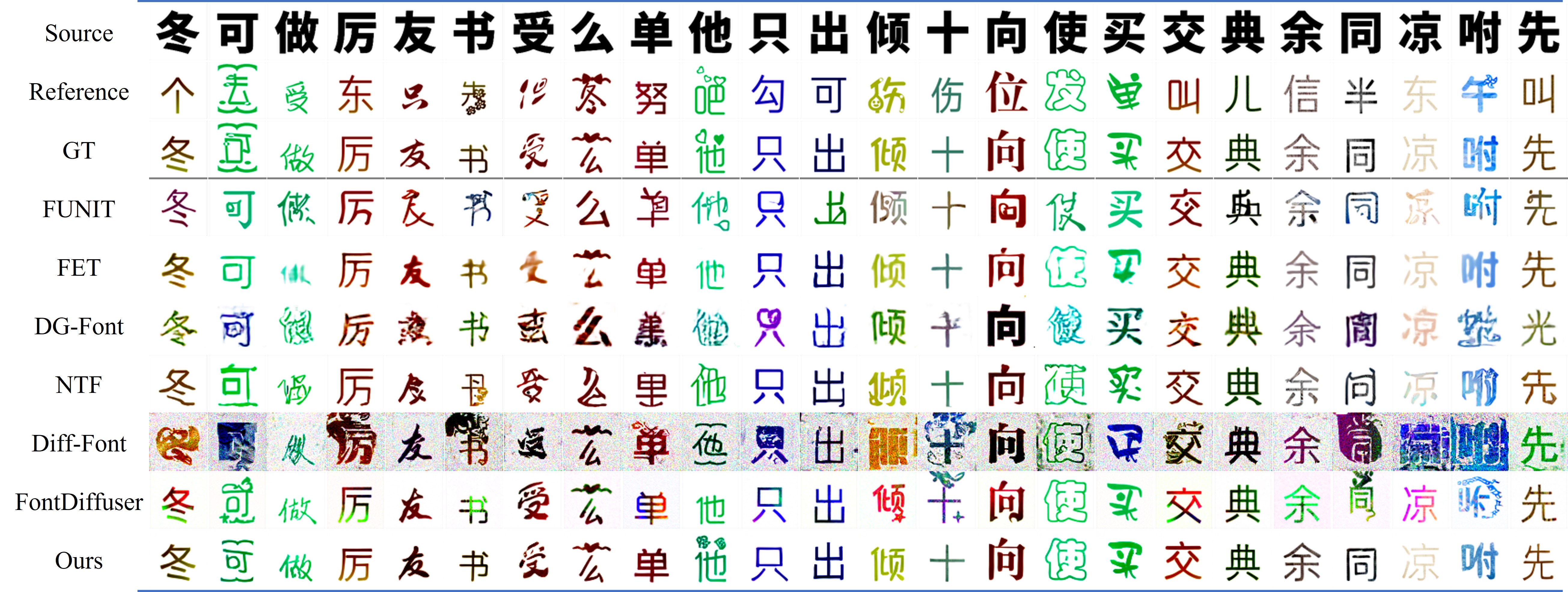}
   \end{overpic}
   \caption{Comparison between competitors and our AnyArtisticGlyph on Chinese100-Dataset}
   \vspace{-10pt}
   \label{fig-exp-chinese}
\end{figure*}

\begin{figure*}
   \begin{overpic}[width=1.95\columnwidth]{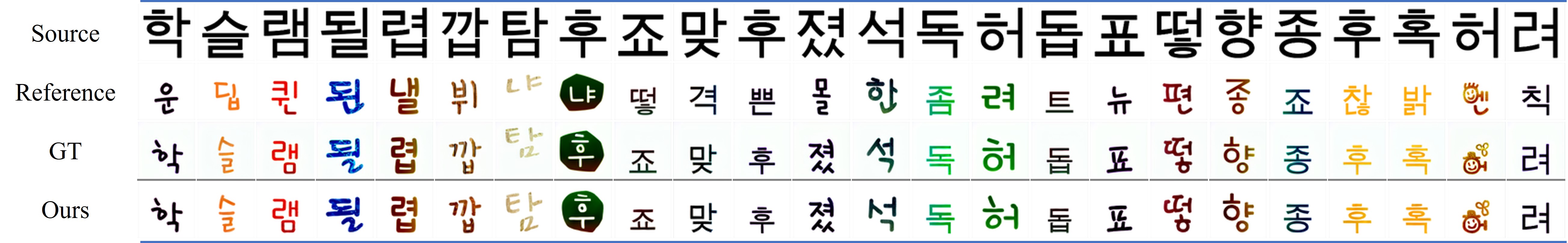}
   \end{overpic}
   \caption{Illustration of generating glyph image on Korean480-Dataset.}
   \vspace{-10pt}
   \label{fig-exp-korean}
   \vspace{-3pt}
\end{figure*}

\subsection{Comparison Details}
\subsubsection{Quantitative Results}
The quantitative results for the MCGAN-Dataset are summarized in Tab. \ref{tab-exp-mcgan}. From the table, it's evident that our proposed AnyArtisticGlyph achieves superior evaluation scores compared to most existing state-of-the-art methods. Across the different evaluation metrics in the table, our method achieves the highest score in four out of five indicators and secures the second position in the remaining two. Specifically, in the pixel-level evaluation PSNR, the performance of AnyArtisticGlyph is 7.4\% better than the second-ranked FontDiffuser. Regarding image generation quality assessed by FID, AnyArtisticGlyph surpasses FontDiffuser by 8.47 points, nearly a 43.22\% improvement. In terms of deep feature space similarity evaluated by LPIPS, AnyArtisticGlyph also outperforms FontDiffuser by 8.40\%. While in SSIM evaluation, our method lags behind FontDiffuser by only 0.0115. 

Moreover, the quantitative results for the Chinese100-Dataset are summarized in Tab. \ref{tab-exp-chinese100}. From the results, we can observe that AnyArtisticGlyph outperforms competing methods in Chinese artistic glyph image generation by a large margin, in terms of pixel-level accuracy (L1, SSIM, PSNR) and realism (FID, LPIPS). Especially for the two indicators FID and LPIPS, improved the performance by 54.31\% and 9.52\% respectively.

\subsubsection{Qualitative Results}
Figure \ref{exp_mcgan} offers a glimpse into the versatility of our method by showcasing a randomly selected test result. This particular example demonstrates the generation of a comprehensive set of English capital letters, taking the letter "R" as the given reference image. The generated images exhibit remarkable fidelity in terms of both glyph shape and texture, highlighting the strength of our approach. Moving beyond English glyphs, we further demonstrate the generalizability of AnyArtisticGlyph by applying it to non-Latin scripts. Figures \ref{fig-exp-chinese} and \ref{fig-exp-korean} provide compelling evidence of this. In the Chinese glyph example, we observe that intricate details of traditional Chinese characters are faithfully reproduced, capturing the essence of their unique style. Similarly, for Korean glyphs, our method manages to replicate the intricate patterns and strokes that characterize the Korean writing system.

\begin{table}[]
    \centering
    \caption{Quantitative Results on Chinese-100 Dataset.}
    \begin{tabular}{c|cccccc}
    \Xhline{2pt}
    \multirow{2}{*}{Model} & \multicolumn{5}{c}{Metrics} \\
     \cline{2-6}
            & L1↓    & SSIM↑  & PSNR↑  & FID↓   & LPIPS↓         \\
    \Xhline{1pt}
    FUNIT   & 0.1213    & 0.5281    & 10.7947    & 70.2739   & 0.2564         \\
    FET     & 0.1059    & 0.5807   & 11.1779    & 38.0623   & 0.2179         \\
    DG-Font & 0.1160   & 0.5798    & 11.0708    & 92.3365   & 0.2720          \\
    NTF     & 0.1062   & 0.5968    & 11.4789    & \underline{32.2743}   & \underline{0.1976}          \\
    Diff-Font  & 0.2781 & 0.2982   & 7.7314    & 110.9636  & 0.4866         \\
    FontDiffuser & \underline{0.1029}  & \underline{0.6310}  & \textbf{12.4482}    & 39.7522   & 0.2144         \\
    Ours    & \textbf{0.1004}     & \textbf{0.6338}    & \underline{12.1465}    & \textbf{18.1636}   & \textbf{0.1940}          \\
    \Xhline{2pt}
    \end{tabular}
    \label{tab-exp-chinese100}
    \vspace{-10pt}
 \end{table}

\subsection{Ablation Study}
% \begin{table*}[]
%    \centering
%    \caption{Ablation experiments of AnyArtisticGlyph on Korean480-Dataset. The results validate the effectiveness of each submodule in AnyArtisticGlyph.}
%        \begin{tabular}{c|cccc|ccccc}
%        \Xhline{2pt}
%        Exp. No. & $c_i$   & $c_t$  &$L_{cx}$  & $\lambda$ & L1↓  & SSIM↑  & PSNR↑  & FID↓  & LPIPS↓          \\
%        \Xhline{1pt}
%    1 & $\times$  & $\times$  & $\times$  & - & 0.0549 & 0.7876   & 17.1658  & 28.8053  & 0.1391    \\
%    2 & $\checkmark$  & $\times$  & $\times$  & - & 0.0535  & 0.7997  & 17.2207 & 26.9131  & 0.1237          \\
%    3 & $\times$  & $\checkmark$  & $\times$  & - & 0.0523  & 0.8062  & 17.8301  & 23.3779  & 0.1333         \\
%    4 & $\checkmark$  & $\checkmark$  & $\times$  & - & 0.0508  & 0.8373 & 18.6664  & 19.3956  & 0.1135          \\
%    5 & $\checkmark$  & $\checkmark$  & $\checkmark$  & 0.5 & 0.0512  & 0.8478 & 18.7613  & 13.7325   & 0.1013          \\
%    6 & $\checkmark$  & $\checkmark$  & $\checkmark$  & 1.0 & \textbf{0.0481}  & \textbf{0.8520} & \textbf{18.8352}   & \textbf{12.2846}  & 0.0808   \\
%    7 & $\checkmark$  & $\checkmark$  & $\checkmark$  & 1.5 & 0.0514  & 0.8452 & 18.4252   & 15.5378  & \textbf{0.0735}          \\
%        \Xhline{2pt}
%        \end{tabular}
%        \label{tab-ablation}
%        \vspace{-10pt}
%    \end{table*}

   \begin{table}[htbp]
    \centering
    \caption{Ablation experiments of AnyArtisticGlyph.}
        \begin{tabular}{p{0.001\textwidth}p{0.001\textwidth}p{0.01\textwidth}p{0.02\textwidth}|ccccc}
        \Xhline{2pt}
        $c_i$   & $c_t$  &$L_{cx}$  & $\lambda$ & L1↓  & SSIM↑  & PSNR↑  & FID↓  & LPIPS↓          \\
        \Xhline{1pt}
    $\times$  & $\times$  & $\times$  & - & 0.0549 & 0.7876   & 17.1658  & 28.8053  & 0.1391    \\
    $\checkmark$  & $\times$  & $\times$  & - & 0.0535  & 0.7997  & 17.2207 & 26.9131  & 0.1237          \\
    $\times$  & $\checkmark$  & $\times$  & - & 0.0523  & 0.8062  & 17.8301  & 23.3779  & 0.1333         \\
    $\checkmark$  & $\checkmark$  & $\times$  & - & 0.0508  & 0.8373 & 18.6664  & 19.3956  & 0.1135          \\
    $\checkmark$  & $\checkmark$  & $\checkmark$  & 0.5 & 0.0512  & 0.8478 & 18.7613  & 13.7325   & 0.1013          \\
    $\checkmark$  & $\checkmark$  & $\checkmark$  & 1.0 & \textbf{0.0481}  & \textbf{0.8520} & \textbf{18.8352}   & \textbf{12.2846}  & 0.0808   \\
    $\checkmark$  & $\checkmark$  & $\checkmark$  & 1.5 & 0.0514  & 0.8452 & 18.4252   & 15.5378  & \textbf{0.0735}          \\
    \Xhline{2pt}
    \end{tabular}
    \label{tab-ablation}
    \vspace{-10pt}
    \end{table}

In this section, we perform several ablation studies to analyze the performance of our proposed modules. The experiments are conducted on the Korean480-Dataset. Since this paper focuses on a text-driven method requiring the fusion of image and text features, we establish a baseline model only utilizing the FFEM. We separate the coarse-grained feature-level loss $L_{fl}$, the conditions $c_i$ and $c_t$, and progressively add them to the baseline, as shown in Tab. \ref{tab-ablation}.

\textbf{Condition $c_i$}. 
Comparing the baseline (Line 1) and $c_i$ (Line 2), we observed that the model's performance is improved to a certain extent when added to condition $c_i$. Specifically, condition $c_i$ was derived using a transformer-based CLIP image encoder, which is renowned for its ability to capture comprehensive visual representations. At this stage, the encoder extracted the global features of the reference image $l_r$, encapsulating vital information about the overall context and structure. These global features, when incorporated into the model, underwent a fusion process with the features of the source image through cross-attention mechanisms. This fusion allowed the model to leverage both the specific details of the source image and the broader context provided by the reference image, leading to an improvement in the model's overall performance.

\textbf{Condition $c_t$}. Line1 and Line3, as well as Line2 and Line4 validate the effectiveness of the text condition $c_t$. Moreover, we observe significant enhancements in perceptual-level evaluation metrics such as FID and LPIPS. The reason behind this is that text prompts play a crucial role in transforming implicit feature learning and transfer into explicit ones. With the introduction of prompts, the model can leverage this information as explicit guidance during the learning process so the model's learning burden is significantly reduced. This not only saves computational resources but also makes the model learning and training process more directional and effective.

\textbf{Loss $L_{fl}$}. 
Line4-7 validate the effectiveness of coarse-grained feature-level loss $L_{cx}$. Upon conducting later three experiments, we found that $\lambda=1$ yielded the best results, with a 10.54\% improvement compared to Line4 in the FID metric. It is worth noting that as $\lambda$ increases, the LPIPS index is still optimized, but other indexes decrease to varying degrees.

\section{Conclusion and Limitations} \label{Limitations}
In this paper, we propose a novel approach called AnyArtisticGlyph, which is a diffusion-based multi-lingual glyph generation framework. Our approach incorporates FFEM that combines text glyphs and reference images into a latent space. Furthermore, the VTFEM uses the CLIP model for encoding reference data as embeddings, which blend with explicit transformation caption embeddings to accomplish seamless global image generation. To improve glyphs accuracy, we employ traditional diffusion loss and coarse-grained feature-level loss during training. Extensive experiments demonstrate AnyArtisticGlyph's superiority over state-of-the-art image-based methods in AGIG. Moreover, experiments on phonetic glyphs and ideographic images confirm the effectiveness of this simple prior-independent design in multi-linguistic generation scenarios. Moving forward, our future work will focus on exploring the generation of unseen content glyph images.

\bibliographystyle{IEEEbib}
\bibliography{icme2025references}
\end{document}